\documentclass[final]{cvpr}

\usepackage{times}
\usepackage{epsfig}
\usepackage{graphicx}
\usepackage{amsmath}
\usepackage{amssymb}

\usepackage[pagebackref=true,breaklinks=true,colorlinks,bookmarks=false]{hyperref}

\def\cvprPaperID{11} 
\def\confYear{CVPR 2021}
\newcommand{\norm}[1]{\left\lVert#1\right\rVert}

\usepackage{soul}
\usepackage[dvipsnames]{xcolor}

\begin{document}

\title{Contrastive Learning for Sports Video:  Unsupervised Player Classification}

\author{Maria Koshkina\\
York University\\
Toronto, Canada\\
{\tt\small koshkina@yorku.ca}
\and
Hemanth Pidaparthy\\
York University\\
Toronto, Canada\\
{\tt\small phemanth@yorku.ca}
\and
James H. Elder\\
York University\\
Toronto, Canada\\
{\tt\small jelder@yorku.ca}
}

\maketitle

\begin{abstract}
We address the problem of unsupervised classification of players in a team sport according to their team affiliation, when jersey colours and design are not known a priori.
We adopt a contrastive learning approach in which an embedding network learns to maximize the distance between representations of players on different teams relative to players on the same team, in a purely unsupervised fashion, without any labelled data. We evaluate the approach using a new hockey dataset and find that it outperforms prior unsupervised approaches by a substantial margin, particularly for real-time application when only a small number of frames are available for unsupervised learning before team assignments must be made.  Remarkably, we show that our contrastive method achieves 94\% accuracy after unsupervised training on only a single frame, with accuracy rising to 97\% within 500 frames (17 seconds of game time).  We further demonstrate how accurate team classification allows accurate team-conditional heat maps of player positioning to be computed.  
\end{abstract}

\section{Introduction}

Team membership classification (i.e. labelling each person on a playing surface as a member of team A, team B or a referee) is a critical task in sports video analytics:  most inferences and statistics depend upon knowing which player are on each team, including attempts on goal, offsides, and player configurations.   Accurate team affiliation labels can also improve player tracking. The problem can be challenging due to the extreme variations in player pose, occlusions, motion blur and uneven illumination. 

Prior work  (e.g., \cite{lu2018lightweight,istasse2019associative}) has framed the problem as a supervised learning task in which labelled data (e.g., bounding boxes with team identifiers) are used to learn a classifier.  Early supervised methods employed hand-crafted colour-based features 
\cite{lu2013learning,liu2014detecting}, while more recent approaches train convolutional neural networks (CNNs) on labelled datasets to perform player segmentation~\cite{istasse2019associative} and classification~\cite{lu2018lightweight}. 
 
Unfortunately, the supervised player classification approach ~\cite{lu2018lightweight} has limited application, since it requires fine-tuning on every new game for optimal classifier performance . The team segmentation approach ~\cite{istasse2019associative} has been found to generalize better but does not provide player instance segmentation and requires expensive pixel-wise annotation to train the system. For all of these reasons, an unsupervised approach is preferred.

To date, unsupervised approaches \cite{mazzeo2010football,ivankovic2014automatic,d2009investigation,bialkowski2014representing,tong2011automatic} rely solely on 
colour-based features such as colour histograms.  While these are simple and lightweight, typically many frames are needed from each new game in order to learn the colour distributions, and these methods fail when the two teams are wearing similar colours.

Our goal in this paper is to understand whether a more powerful representation, that may include both colour and configural information, can be learned in a fully unsupervised manner, and whether such a representation can reduce the number of frames needed for training and improve generallization to novel teams, jerseys, lighting and camera parameters.

\begin{figure*}
\begin{center}
\includegraphics[width=0.8\linewidth]{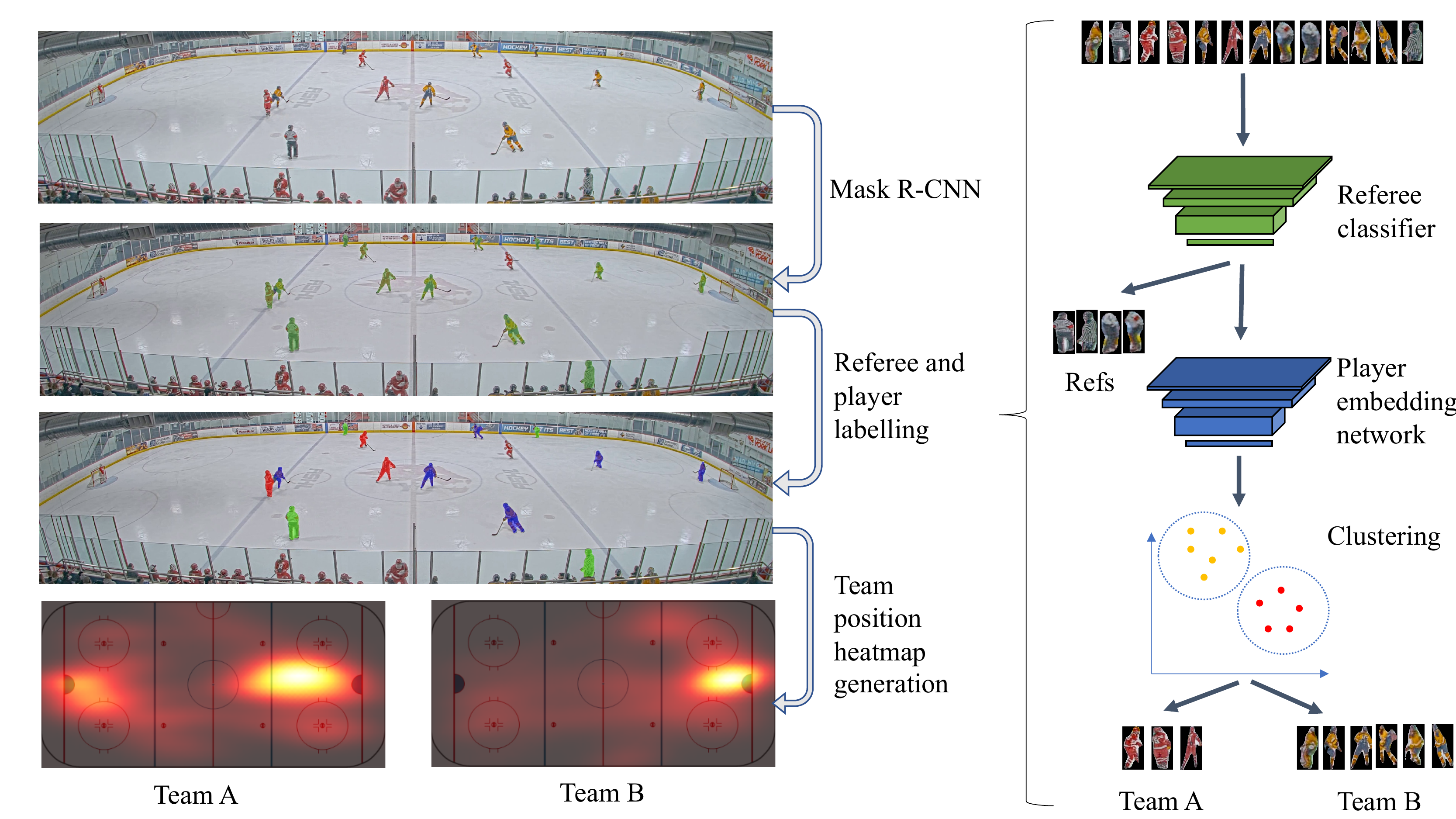}
\end{center}
   \caption{Overview of the proposed system.  Mask R-CNN is first used to detect and segment each person on the playing surface.  A pre-trained CNN is then used to classify referees, whlie remaining players are passed to our embedding network for clustering into teams.  This allows production of heat maps showing the distribution of the two teams over the playing surface.}
\label{fig:workflow}
\end{figure*}

To achieve this, we employ unsupervised contrastive learning to train a CNN to cluster players into two teams.     We demonstrate our system's performance on a new hockey dataset and compare it to previously proposed unsupervised team affiliation learning approaches.   Figure~\ref{fig:workflow} demonstrates overall system design.
The dataset and code are available at \href{https://github.com/mkoshkina/teamId}{https://github.com/mkoshkina/teamId}.

Our main contributions are:
\begin{enumerate}
\item We introduce what is, to our knowledge, the first unsupervised deep learning approach for team classification.  This novel contrastive learning approach allows us to generalize to novel games, teams and jerseys without labelled data.
\item We introduce a new annotated hockey dataset that can be used to evaluate player detection and team classification algorithms.
\item We show that our novel unsupervised algorithm outperforms prior unsupervised approaches by a large margin, especially when only a small number of frames are available for unsupervised learning before team assignments must be made.  This limits the burn-in time for real-time streaming applications and allows the system to adapt quickly to changes in lighting or camera parameters.
\item We show  how our system for team classification can be used to produce accurate team-conditioned heat maps of player positioning, useful for coaching and strategic analysis.
\end{enumerate}

\section{Related Work}

\subsection{Player Classification} Automatic labelling of players according to team is critical for sport video understanding, including player tracking \cite{lu2013learning,liu2014detecting,tong2011automatic,shtrit2011}, player configuration analysis, activity recognition \cite{bialkowski2014representing}  and detection of game events \cite{d2009investigation}.

Early work relied on colour histograms \cite{mazzeo2010football,shtrit2011,ivankovic2014automatic,d2009investigation,lu2013learning,liu2014detecting,bialkowski2014representing}  and `bag of words' representations of  colour features \cite{tong2011automatic}. These approaches are lightweight, however the exclusive reliance on colour features make them more sensitive to illumination changes and could lead to lower performance when teams are wearing similar colours.  

In recent years, supervised deep learning based methods for player detection and player labelling have been proposed \cite{lu2018lightweight, istasse2019associative}. These methods perform well but require labelled data for training. In \cite{istasse2019associative}, a  CNN is trained to segment players and generate team pixel-wise descriptors, where pixels of players from the same team have descriptors that are close in embedding space. Pixels are then clustered to identify the players on the two teams. This method requires pixel-level team labelling to train the network and per-image pixel-level clustering at the inference stage.  Moreover, it does not provide instance-level segmentation so would not be suitable for use in player location heatmap generation.

Lu {et al.}~\cite{lu2018lightweight} also take a supervised approach, employing a cascaded CNN to learn team membership classification (team A, team B and others) from labelled data.  This method has good results but does not generalize well and thus requires fine-tuning on labelled samples from a new game in order to be used for that game.

Clearly, both simple colour-based unsupervised approaches and more sophisticated CNN-based supervised approaches have limitations.  Here we explore whether modern deep unsupervised learning methods can be used to overcome these limitations.

\subsection{Contrastive Learning and Deep Clustering}
Contrastive learning~\cite{hadsell2006dimensionality}  is a self-supervised representation learning approach that aims to map similar objects to be close in embedding space and dissimilar objects further apart, and has been shown to produce excellent results on a number of tasks \cite{he2019momentum,chen2020simple}.  In our work we use a simple CNN trained with triplet loss \cite{hoffer2015deep} to learn a feature space that best separates players into two teams.  

Recent work in contrastive learning \cite{he2019momentum,chen2020simple}  shows excellent results in unsupervised representation learning on large datasets such as ImageNet\cite{ILSVRC15} or COCO\cite{lin2014microsoft}.  These methods are based on noise contrastive estimation and involve using an anchor (typically an augmented version of an original image), one positive (another augmented version of the same image) and a large number of negatives, randomly picked from the training set.  This setup works well for a dataset with a large number of categories, where randomly-picked images are unlikely to contain many positives. In our setting however, although we have a large number of images we have a relatively small number of categories (unique jersey designs).  More precisely, ImageNet contains 1000 categories and our training dataset has 10.   As a consequence, in our setting using random images as negatives results in a 10\% of false negatives.  This adversely affects training. For this reason, a simple triplet loss  works much better in our setting.

Our work is  inspired in part by deep clustering approaches  \cite{yang2017towards,yang2016joint} in which CNNs are used to jointly learn feature representations and cluster centres in an unsupervised fashion.  In our approach we use pseudo-labels from an initial k-means clustering as a supervision signal to train our contrastive learning CNN.  The main divergence from prior methods is that we are only interested in learning feature space that will lead to good data separation - cluster centres can be quite different in each new game.  Once trained, the network is only used to extract features from player images.  

\section{Method}
\subsection{Overview}Our general goal is to develop automatic sports  analysis tools that provide valuable visualizations, statistics and analyses for coaches and players. Our current work is focused on hockey, but can easily be adapted to other team sports such as soccer, basketball and football.  In this paper we design and evaluate a system that automatically detects players, classifies them into teams and returns a heatmap of the distribution of players for each of the two teams.   

We employ video from a stationary 4K camera that captures the whole playing surface, and use the Mask R-CNN network~\cite{he2017maskrcnn}) to detect and segment all people on the ice, including the players from the two teams and the referees.  Since the referee uniform is consistent across games, we first train a CNN to perform referee classification based upon labelled data (referee, non-referee).   

In order to classify players we employ an embedding CNN trained with triplet loss to extract a learned feature vector for each player image.  We then use k-means to estimate cluster centres for the two teams from one or more initial frames of the video.  On all subsequent frames we assign each player to a team based on the closest cluster centre in feature space.  Using a learned homography, we geo-locate each detected player on the ice surface and use kernel density estimation (KDE) to construct a heatmap representing the distribution of players across the playing surface for each team.    Figure~\ref{fig:workflow} shows the pipeline for our system.

\subsection{Dataset}
\label{sec:dataset}
We introduce a new labelled hockey video dataset that will be made public on acceptance of this paper.  Despite the variety of available sport video datasets \cite{apidis,giancola2018}, to the best of our knowledge our new hockey dataset will be the only publicly-available sport video dataset that contains team affiliation labels. 

The dataset is drawn from 15 different hockey games captured over two seasons.  Seven of the games (season 1) are captured with a wide-field stationary 4K (3840 $\times$ 2160 pixel) 30 fps camera that captures nearly the whole rink.  In order to better capture the whole rink, season 2 games are captured by two 4K cameras with 75 degree horizontal displacement, together capturing the whole rink with modest overlap.  We defined a virtual camera with intrinsic parameters matching the two real cameras and extrinsic parameters equal to the mean of the two real cameras.   Each of the two camera images was rectified to the virtual camera through a homography with the ice surface. The two virtual images were then smoothly blended. The resulting season 2 videos have resolution of 5930 $\times$ 1080 pixels.

We manually close-cropped the videos to the 3840$\times$900 (season 1 games) and 5680$\times$904 (season 2 games) rectangle bounding the rink (Fig. \ref{fig:workflow}).  From each game we randomly extracted a video clip of roughly 850 frames (28 sec).   Each game contains a unique combination of player uniforms, and since play is active in each clip there is considerable variation in player pose, motion blur and occlusions between players.  Players were automatically detected using Mask R-CNN (see below).  To eliminate coaches and bench players, we applied a heuristic to exclude detections close to the bottom of the frame that had bounding box height less than twice the width.

 {\em For evaluation only}, we manually annotated every 10th frame of each game clip, thus obtaining between 80-90 labelled frames per game.  Annotations consist of:
\begin{enumerate}
\item Mask R-CNN detections
\begin{itemize}
\item Class label (Team A, Team B, Referee)
\end{itemize}
\item Manual detections (including players not detected by Mask R-CNN) 
\begin{itemize}
\item Class label (Team A, Team B)
\item Estimated image projections of points of contact with playing surface (skates on the ice)
\end{itemize}
\end{enumerate}

To label the Mask R-CNN detections, extracted player images with segmentation masks applied were manually inspected one-by-one. Only the images that could be identified by visual inspection to belong to a player (Team A, Team B ) or referee were labelled, the rest were marked as false positives.
If there were multiple players within one extracted image the player with the most pixels in the mask was selected. 

We use the Mask-RCNN labels of detected players to evaluate the accuracy of team classification algorithms and use the manual detection annotations to evaluate the accuracy of our team positioning heatmaps.

The 15-game dataset was divided into training, validation and test sets with a 9-2-4 split.  Both training and test set contains a mix of season 1 and season 2 games. One limitation of the dataset is that even though each game has a unique combinations of teams playing, some teams appear multiple times through the dataset.  We have ensured that the test set includes a game with previously unseen teams.

\subsection{Player Detection and Segmentation}
\label{sec:detection}
We employ Mask R-CNN \cite{he2017maskrcnn} trained on MS COCO \cite{lin2014microsoft} to detect and segment all people on the playing surface.  To adapt to the different resolution, aspect ratio and expected size of people in our video relative to MS COCO, we partitioned each frame into left and right  images with a 40 pixel central overlap, running Mask R-CNN on each individually before merging results.  Bounding boxes detected in the left image that overlap boxes detected in the right image by 45\% or more are merged by selecting the larger of the two boxes.   We define the estimated image location of each player as the mid-point of the lower boundary of the R-CNN bounding box. 

\subsection{Referee Classifier}
\label{sec:classifier}
Since referee uniforms are consistent across games, a supervised approach is appropriate.  We use the referee/non-referee labels from our Mask R-CNN detections to train and evaluate a simple CNN classifier. This is the only way that labelled data is used in our system, aside from evaluation.  Our CNN classifier takes as input R-CNN detection images with segmentation masks applied and classifies them as referee or non-referee.  We employ a small CNN with 3 convolutional layers (16, 32, and 64 output channels)  and 3x3 kernels followed by 2 fully connected layers.  We train the network with a binary cross entropy loss function, employing the Adam optimizer.   

\subsection{Unsupervised Team Assignment: Feature Learning and Clustering}
\label{sec:features}
An ideal team labelling algorithm will be unsupervised, generalizing to new games without needing any labelled data, and will require minimal frames (burn-in time)  from the beginning of the game to determine accurate labels for each player on the team. 

Previous unsupervised approaches used colour features such as histograms and bag-of-colours.  These approaches can be effective but since they do not consider spatial features, performance may suffer when teams are wearing jerseys with similar colour profiles, or when illumination variations render colour features unreliable.  Here we explore whether an embedding CNN trained by contrastive learning can produce a more powerful representation that, by incorporating both colour and spatial features, can learn a reliable feature representation from fewer frames, and thus have a shorter burn-in time.  

We employ a CNN with 3 convolutional layers (16, 32, and 64 channels) and 3x3 kernels, each followed by a pooling layer, and two fully connected layers. The last layer returns a feature vector of length 1024. We train our network using the Adam optimizer on a training set of games using a triplet loss \cite{hoffer2015deep}. Input is a triplet of extracted images with the R-CNN mask applied: an anchor image, a positive image and a negative image.  The positive image is an image of a player believed to be from the same team as the anchor image, while the negative image is a player believed to be on the other team.  The triplet loss function, when back-propagated, drives the network to decrease the distance in the embedding space between the anchor and positive images, while increasing the distance between the anchor and negative images.  In order to ensure that the learned representation does not exclusively rely on colour, we randomly convert 50\% of training triplets to grayscale.

Unsupervised training of the embedding network requires a method for estimating whether two input images have the same or different labels.  We seed this process with a simple colour-based distance measure, representing each image as a normalized RGB histogram with 8 bins per colour channel  and then using k-means  to cluster players into two teams.  

To form the triplets, we  first rank the player images $\mathbf{x}_i$ in terms of their team assignment confidence scores $p_{ij}$, using a standard
`soft k-means' measure:
\begin{eqnarray}
p_{i1} &=& \frac{\norm{\mathbf{x}_i-\mathbf{c}_2}}{\norm{\mathbf{x}_i-\mathbf{c}_1}+ \norm{\mathbf{x}_i-\mathbf{c}_2}}\\
p_{i2} &=& \frac{\norm{\mathbf{x}_i-\mathbf{c}_1}}{\norm{\mathbf{x}_i-\mathbf{c}_1}+ \norm{\mathbf{x}_i-\mathbf{c}_2}}
\end{eqnarray}
where $\mathbf{c}_j$ is the centre of cluster $j$ and $p_{ij}$ is the confidence with which image $i$ is assigned to cluster $j$.

 We consider only high-confidence samples ($p_{ij}>0.9$)) for training to limit the label noise.  We then randomly form triplets by sampling the anchor and positive images from one cluster (anchor and positive example) and the negative image from the other.  
  
\begin{figure*}
\begin{center}
\includegraphics[width=0.8\linewidth]{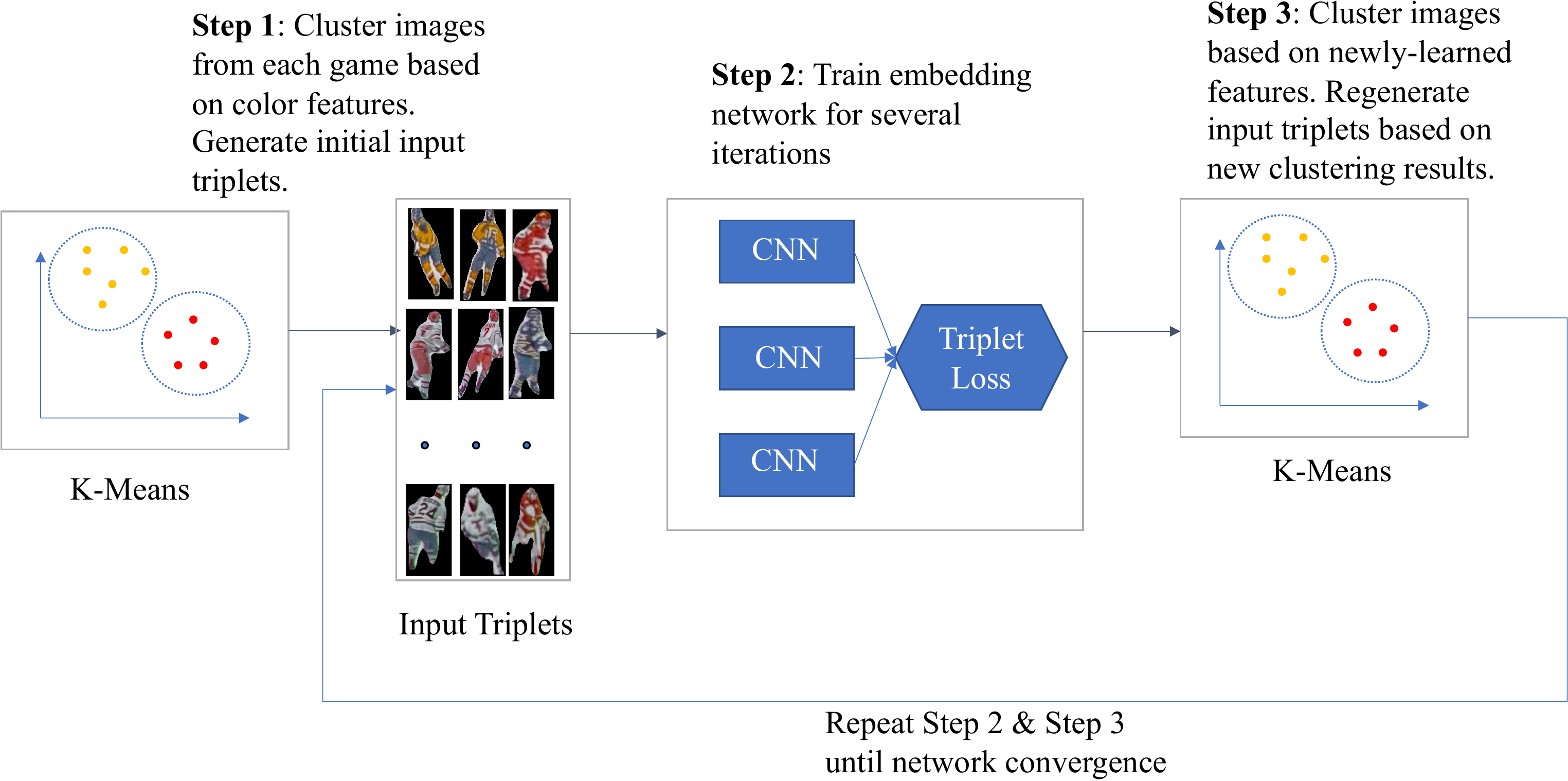}
\end{center}
   \caption{Self-supervised training of embedding network.}
\label{fig:network}
\end{figure*}

As the training proceeds we regenerate these pseudo-labels and training triplets, but replacing the histogram representation with the evolving embedded representation learned by the network.  We train until convergence (no improvement on the validation data for 3 epochs) or a maximum of 30 epochs on the initial colour histogram pseudo-labels and then generate new pseudo-labels from the evolving embedded representation every 10 epochs (or until convergence).  We find that the proportion of high-confidence samples  grows over time, indicating that the network is learning a representation that improves data separation. Figure~\ref{fig:network} illustrates this training process. 

\begin{figure*}
\begin{center}
\includegraphics[width=0.7\linewidth]{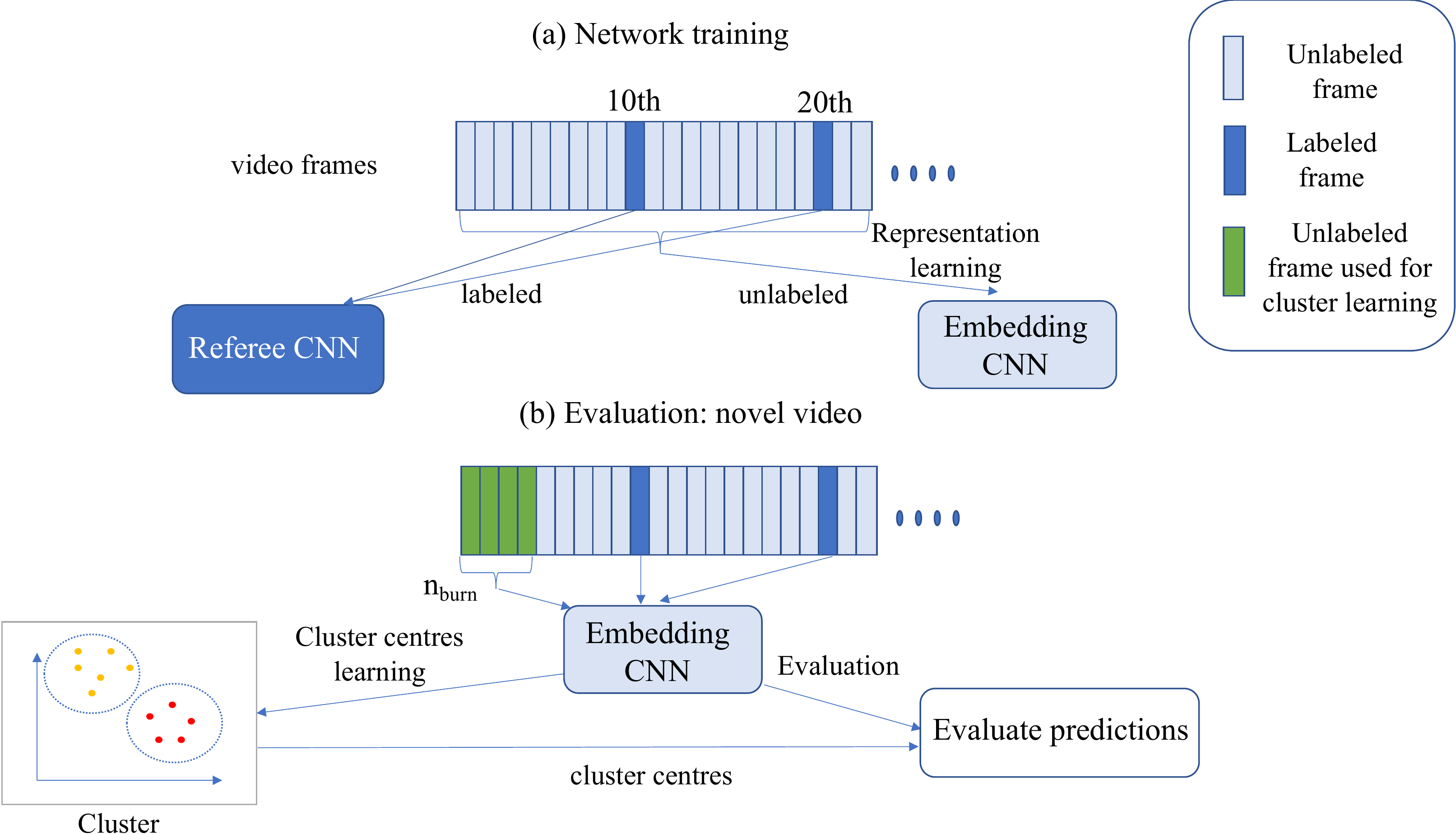}
\end{center}
   \caption{Data usage for training and evaluation.  a) Labelled frames are used to train the referee classifier, but the embedding network representation is learned in unsupervised fashion, without reference to labels.   b) To perform inference on a novel video, we use the first $n_{burn}$ frames to find cluster centres in the learned embedded representation.  Labelled frames are used only for  evaluation. }
\label{fig:data_usage}
\end{figure*}

Once unsupervised training of the embedding network on the training set is complete, we apply the network to novel games with unseen teams and uniforms.  We use the first $n_{burn}$ frames of the unseen game as input to k-means to determine the two cluster centres for this new game in the pre-learned embedding space.  Once the cluster centres are identified, we associate detected players in subsequent frames with the nearest cluster centre, and evaluate on annotated frames.    Figure~\ref{fig:data_usage} illustrates our use of data for training and evaluation purposes.

\subsection{Team Positioning Heatmaps}
\label{sec:heatmaps}
One of the many useful applications of player detection and labelling is the generation of team positioning heatmaps that can help coaches and players understand how their players and the players on the opposing team tend to be distributed throughout a game or portion of the game. 


To generate these heatmaps we use a learned homography to transfer the image coordinates of each detected player (midpoint of the bottom of each bounding box) to the corresponding point on a model of the playing surface.   The homography was computed from 19 corresponding pairs of points in one video frame and in a template model of the ice rink (Illustration showing keypoints and backprojected player positions is included in supplementary material).  We then used the standard least-squares reprojection method \cite{luong1996fundamental} to estimate the homography mapping image pixels to points on the ice surface.   Based on these player positions and the team affiliations estimated using our unsupervised contrastive learning algorithm over multiple frames, we compute a rectified map of player density (players per square metre per frame) using Gaussian kernel density estimation (KDE) \cite{rosenblatt1956, parzen1962}.   Figure~\ref{fig:heatmaps} shows examples of these automatically-generated maps.

\section{Evaluation}

\subsection{Implementation Details}
Our system is implemented in Python 3 with Pytorch and Sklearn.  We use the publicly-available Mask R-CNN network and weights \cite{matterport_maskrcnn_2017} with a confidence threshold of 0.6.   Both referee and embedding networks take as input player images with segmentation masks applied, resized to 62$\times$128 pixels, roughly the average size of a player image.  To reduce the impact of illumination variations we applied an affine transform $I^\prime _i(x,y) = aI(x,y)+b$ to the intensities $I_i$ of all three channels $i\in\{R,G,B\}$ such that $\min_{x,y,i}I^\prime_i = 0$ and $\max_{x,y,i}I^\prime_i = 255$. 

K-means computation of cluster centres entails 10 random initializations:  The solution that minimizes the mean squared deviation from cluster centres is selected. 

\subsection{Comparison with Other Unsupervised Approaches}
\label{sec:results}
We compare the performance of our unsupervised team affiliation algorithm against the two main previously proposed unsupervised team labelling approaches:   colour histograms~\cite{mazzeo2010football,ivankovic2014automatic,d2009investigation,lu2013learning,liu2014detecting,bialkowski2014representing}  and bag-of-words representations of  colour features~\cite{tong2011automatic}.  Since the code and datasets for these previous approaches are not available, we performed a hyperparameter search using k-fold cross-validation to determine the optimal parameters and use k-means clustering to determine cluster centres.  These optimal parameters were the number of bins per channel for the histogram algorithm and number of words for the bag of colours algorithm.  In addition, we evaluated whether to use the entire segmented player or just the upper half, since  the lower half of the uniform is fairly consistent across teams, and also experimented with multiple colour spaces (see below).   

We also experiment with replacing features learned by our contrastive learning network with features learned with convolutional autoencoder (see Section~\ref{sec:ae}).

Comparison with previously used supervised approaches \cite{lu2018lightweight, istasse2019associative} is not feasible as the code and datasets are not available.

\subsubsection{Colour Histogram Algorithm}
Our colour histogram method simply histograms the colours within the segmented player, normalizing by the number of pixels.
We experimented with RGB, LAB and HSV colour spaces, and also tried eliminating the luma or value channel (i.e., two-dimensional AB and HV spaces) to reduce sensitivity to illumination, but found optimal performance with RGB coding.

Cluster centres are then found using k-means, using Euclidean distance in the colour histogram space.  The single hyperparameter is the number $n$ of bins per channel:  k-fold cross-validation revealed that $n=8$ produces best results for our dataset.  We also found that performance was slightly better if only the upper half of the segmented player was considered as the player jerseys are most distinct between teams.

\subsubsection{Bag-of-Colours Algorithm}
\label{sec:bag}
In our bag-of-colours method, we employ the expectation maximization algorithm to fit a Gaussian mixture model (GMM) with $n$ components to the normalized colours of the players in the initial training partition of the novel game.  These components then form the words of a dictionary with which to encode players in subsequent frames.  K-fold cross-validation reveals that $n=35$ components yields optimal results.  We  use k-means clustering to find each team's cluster centres in this 35-dimensional space and assign  players to the closest cluster.  We again find that considering only the top half of the segmented player yields superior results. We also consider a variation of this approach, pretrainied bag-of-colours,  where the dictionary of colours is learned on training set games.

\subsubsection{Autoencoder}
\label{sec:ae}
For additional comparison, we use small convolutional autoencoder \cite{masci2011stacked} trained on image reconstruction.  The encoder network architecture is kept the same as our embedding network and decoder mirrors the encoder.  After training on the images in our training set, we use encoder portion to extract 1024 feature vector for each test image.  We then use these features in the same setting as our embedding features to first learn cluster centres on the burn-in frames and then assign players to closest centre for the rest of the frames.

\subsection{Evaluation Methodology}
We evaluate team affiliation labelling on players detected by mask R-CNN.  These include false positives and imperfect segmentations.  In addition, since the referee classifier is also imperfect, some referees will be incorrect classified as players and will add noise to the contrastive learning process.   We test both our supervised referee classifier and our unsupervised embedding network team classifier on the test set consisting of 4 games.  

Accuracy is evaluated over the 30 annotated frames immediately following the burn-in interval.  Since every tenth frame is annotated, this represents roughly 10 sec of video at 30 fps.

We assess effects of noise in initial pseudo-labels on embedding network performance by considering different team assignment confidence scores $p_{ij}$ thresholds.  Higher confidence threshold leads to better clustering performance. We include this evaluation in supplementary materials . 

\subsection{Referee Classification}
For each game we have 80-90 frames annotated frames and there are 3-4 referees on the rink, so for 9 training games we have 2000 referees in our training set, augmenting these by left/right reflections yields a total of 4000 training vectors.  Employing a softmax threshold of 0.5, we achieve a mean accuracy of 98\%  with 93\%, precision, 96\% recall.  Precision-recall curve for referee classifier is included in supplementary materials.


\subsection{Team Classification}
Table~\ref{table:results} shows the mean accuracy of team classification for the all algorithms under evaluation.  Results depend upon the number $n_{burn}$ of frames  available for learning cluster centres prior to inference.  When $n_{burn}$ is large (512 in this case), two colour-based and our methods perform fairly well, with colour-based methods rivalling our CNN approach.  However, when $n_{burn}$ is small (1 in this case), performance of the colour-only methods drops dramatically, while our embedding CNN approach still performs very well.  

\begin{table}
\begin{center}
\begin{tabular}{|l|c|c|}
\hline
\bf{Method} & $\mathbf{n_{burn}=1}$ & $\mathbf{n_{burn}=512}$ \\
\hline\hline
Colour Histogram & $0.87 \pm  0.031$ & $\boldsymbol{0.97 \pm 0.012}$\\
Bag-of-colours & $0.76 \pm 0.032$ & $\boldsymbol{0.97 \pm  0.018}$\\
Pretrained Bag-of-colours & $0.86 \pm 0.099$ & $0.89 \pm  0.189$\\
Autoencoder & $0.70 \pm 0.076$ & $0.92 \pm  0.099$\\
\bf{Embedding CNN} &  $\boldsymbol{0.94 \pm 0.009}$ & $\boldsymbol{0.97 \pm 0.011}$\\
\hline
\end{tabular}
\end{center}
\caption{Team classification accuracy as a function of the number $n_{burn}$ of frames available for learning cluster centres prior to inference. We show mean and standard error of the accuracy over 4 test games.}
\label{table:results}
\end{table}

\begin{figure}[t]
\begin{center}
   \includegraphics[width=0.9\linewidth]{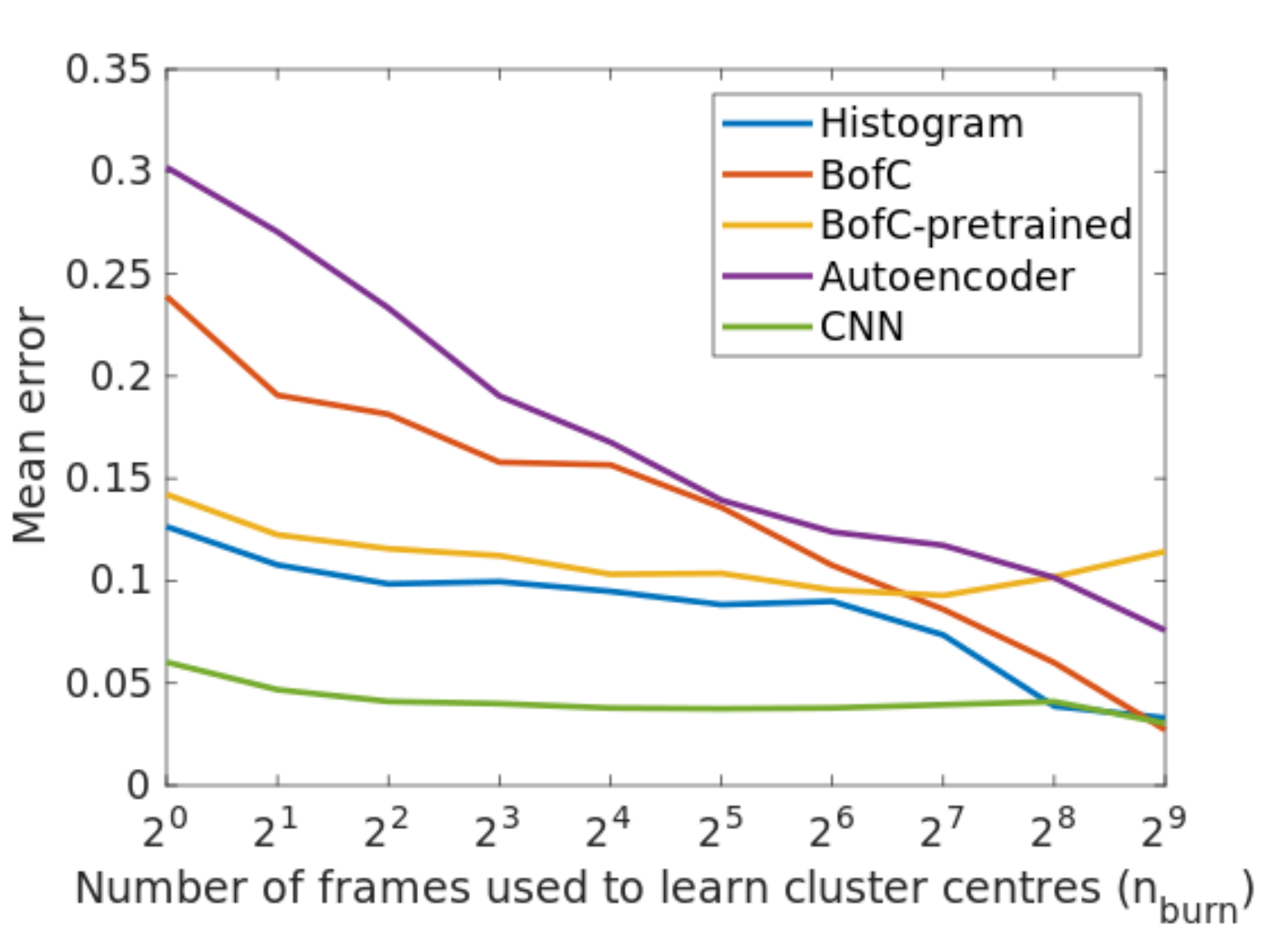}
\end{center}
   \caption{Error rate as a function of the number $n_{burn}$ of initial frames used to learn cluster centres.  We show mean error of the accuracy over 4 test games}
\label{fig:frames_experiments}
\end{figure}

This behaviour is shown in more detail in  Fig.~\ref{fig:frames_experiments}.  We see that the simpler colour-based approaches and autoencoder approach improve continuously as the number of training frames increases, while our embedding CNN approach performs well even with only one training frame, improving only modestly thereafter.  At least 512 burn-in frames are required before the pure colour approaches begin to rival our embedding CNN algorithm.  Autoencoder method is lagging behind even with 512 burn-in frames.

We believe that the advantage of our embedding CNN approach derives from the ability of our unsupervised contrastive learning network to learn from the training games an embedding space that is more effective for discriminating teams than colour histograms.  This more discriminative space then allows well-separated cluster centres to be learned very quickly from the novel game.

\begin{figure*}[h]
\begin{center}
\includegraphics[width=0.8\linewidth]{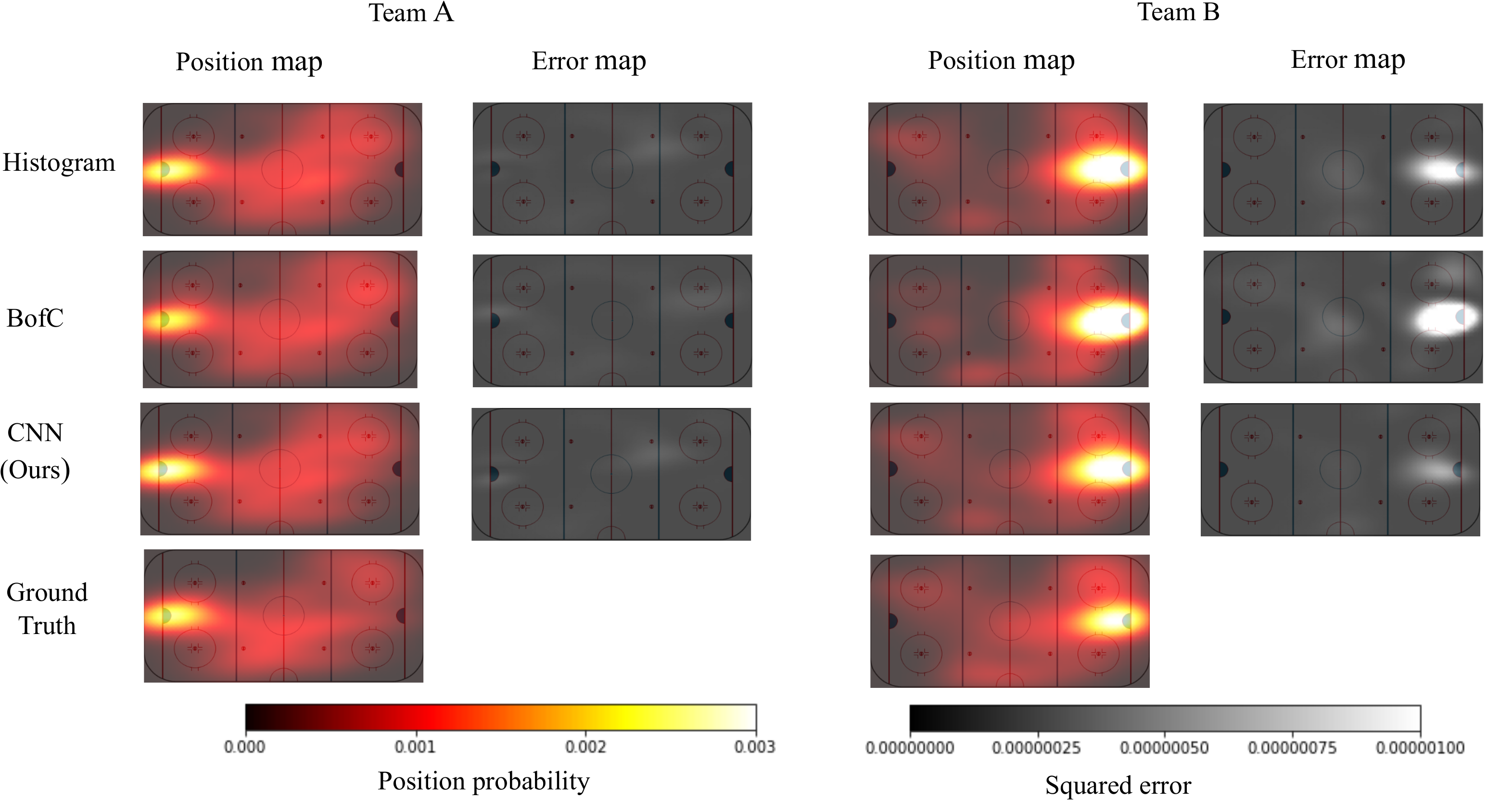}
\end{center}
   \caption{Team positioning heatmaps for a test game.}
\label{fig:heatmaps}
\end{figure*}

\subsection{Team Position Heatmaps Results}
One useful application of player detection and team classification is to allow visualization of team positioning over the course of a game or a portion of a game.  We demonstrate this by generating heatmaps for each game based upon the 800-900 frames used for each game in our experiments.  A learned homography is employed to back-project the image location (midpoint of the bottom boundary of the bounding box) to the playing surface.  We also back-project our manual detections (Section \ref{sec:dataset}) to form a ground-truth heatmap.  Gaussian kernel density estimation~ \cite{rosenblatt1956, parzen1962} is then used to estimated the player density (players per metre squared per frame) for both estimated and ground-truth heatmaps.  The Gaussian bandwidth for KDE is calculated using Silverman's rule of thumb \cite{silverman1986density}, and is roughly 30 pixels for all images (template rink image size is 496x240 pixels).

Fig. \ref{fig:heatmaps} shows example results from one test game for the three team classification methods using $n_{burn}=1$ frames to learn cluster centres.  We see that our embedding CNN approach more consistently represents the true player densities than the pure-colour histogram or bag of colours approaches.

For quantitative evaluation, we scale the maps to integrate to one and then compute the KL-divergence between estimated and ground truth densities over our test set (Table~\ref{table:kl}).  While the bag-of-colours algorithm outperforms a simple colour histogram, our embedding CNN approach substantially outperforms both pure-colour methods.

\begin{table}
\begin{center}
\begin{tabular}{|l|c|}
\hline
{\bf Method} & {\bf Mean KL-divergence} \\
\hline\hline
Colour histogram &0.072 \\
Bag-of-colours & 0.069 \\
{\bf Embedding CNN (Ours)} & {\bf 0.047}\\
\hline
\end{tabular}
\end{center}
\caption{KL-divergence of automatically-generated player positioning heatmaps from ground truth.}
\label{table:kl}
\end{table}

\subsubsection{Runtime}
Our experiments are conducted on 3.6GHz Intel Core i9 CPU x 16 with 64 GB RAM and an Nvidia GeForce RTX 2080 GPU.  Our method runs in real time on segmented player images.  It takes 21 miliseconds to learn team appearances for the game from a single frame and 11 miliseconds per frame for inference on subsequent frames. For convenience, we employed the widely-available but non-real-time Mask R-CNN network~\cite{he2017maskrcnn} for player detection and segmentation, which runs at roughly 5fps.  If replaced with a real-time segmentation network, such as Yolact~\cite{bolya2019yolact}, our whole system will run in real-time.  We leave this for future work.


\section {Conclusions \& Future Work}
Our results demonstrate that a learned representation that can incorporate both colour and spatial features can produce superior results for team classification than a pure-colour approach.  We also demonstrate that such a representation can be learned in unsupervised fashion, using contrastive learning with a triplet loss.  A major benefit is that unsupervised pre-learning of the representation allows for ultra rapid learning of cluster centres from novel games, which limits the burn-in period, allowing online inference.  We also show how this approach to team classification can be used to produce accurate team-conditional player positioning maps that can be useful for coaching and game analysis.  

Improvements could be made by integrating with player tracking:  While team classification will aid tracking,  the converse is also true:  Tracking can potentially
eliminate occasional errors in team classification.  


\textbf{Acknowledgements.} We acknowledge the support of the York VISTA (vista.info.yorku.ca) and Research Chair programs. We thank Canlan Ice Sports (www.canlansports.com) and York University Athletics for providing the video data for this project.

{\small
\bibliographystyle{ieee_fullname}
\bibliography{team_affiliation_cvpr2021}
}

\end{document}


\title{Supplementary Materials\\
Contrastive Learning for Sports Video:  Unsupervised Player Classification}


\maketitle

\section{Effect of Noise in Pseudo-Labels }
To evaluate effects of noise in initial pseudo-labels on embedding network performance we consider different team assignment confidence scores $p_{ij}$ thresholds.  Setting higher threshold results in selecting fewer samples in the initial training set, while using lower threshold results in having a larger but noisier sample set.  Table~\ref{table:noise} shows that selecting higher confidence threshold leads to better clustering performance.

\begin{table}[h]
\begin{center}
\begin{tabular}{|l|c|c|}
\hline
\bf{Threshold} & \bf{Error Rate} & \bf{ Train Set Size} \\
\hline\hline
$p_{ij} > 0.5$  &0.134& 77\%\\
$p_{ij} > 0.7$  &0.037 & 72\%\\
$p_{ij} > 0.9$  &0.031 & 60\%\\
\hline
\end{tabular}
\end{center}
\caption{Team classification error as a function of initial pseudo-labels confidence threshold.  We indicate pecentage of training samples that satisfied the threshold and were included in the initial training set. }
\label{table:noise}
\end{table}

\section{Homography}
\begin{figure*}[h]
\begin{center}
\includegraphics[width=0.9\linewidth]{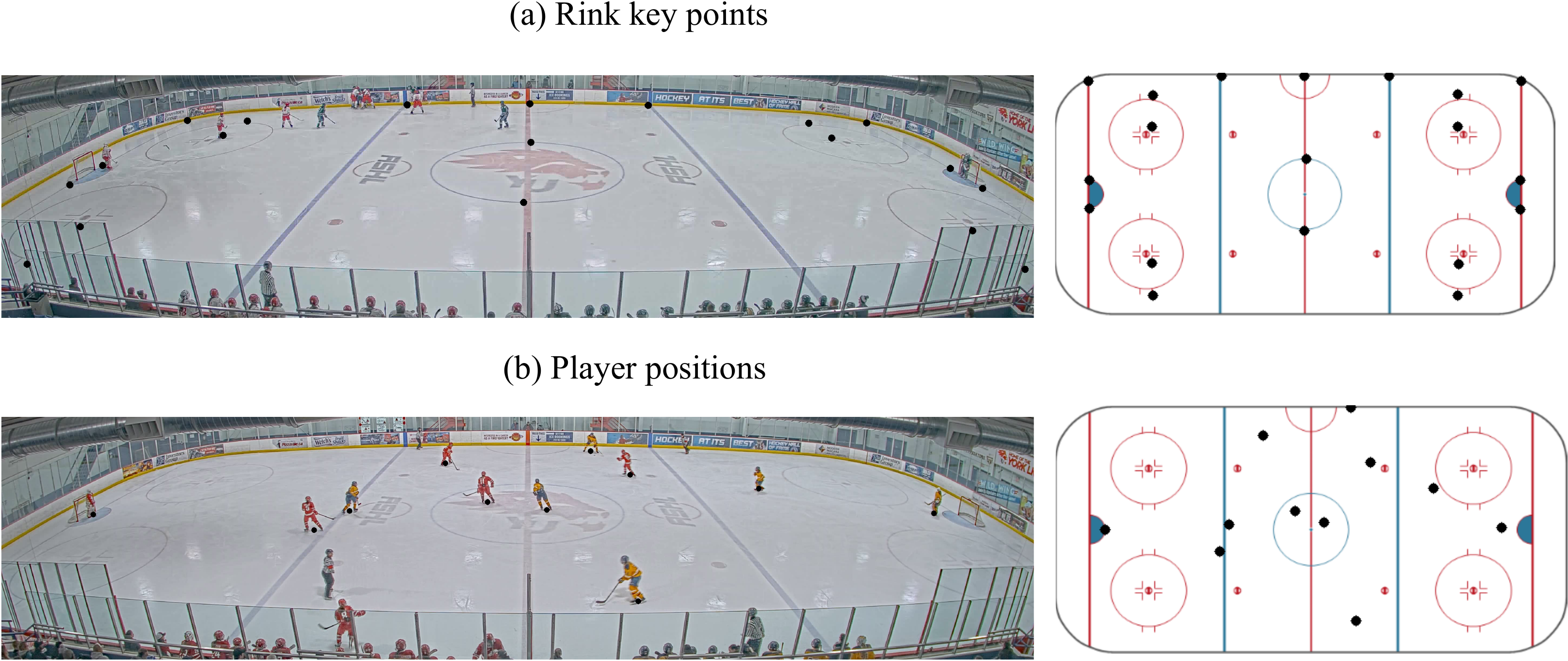}
\end{center}
  \caption{Homography mapping video pixels to ice coordinates. (a) Keypoint pairs.  (b) Backprojected player positions.}
\label{fig:homography}
\end{figure*}

In order to generate team position heatmaps, we use a learned homography to transfer the image coordinates of each detected player (midpoint of the bottom of each bounding box) to the corresponding point on a model of the playing surface.   The homography was computed from 19 corresponding pairs of points in one video frame and in a template model of the ice rink Figure~\ref{fig:homography} shows keypoints and backprojected player positions.
 
\section{Referee Classifier}
\begin{figure}[H]
\begin{center}
   \includegraphics[width=0.9\linewidth]{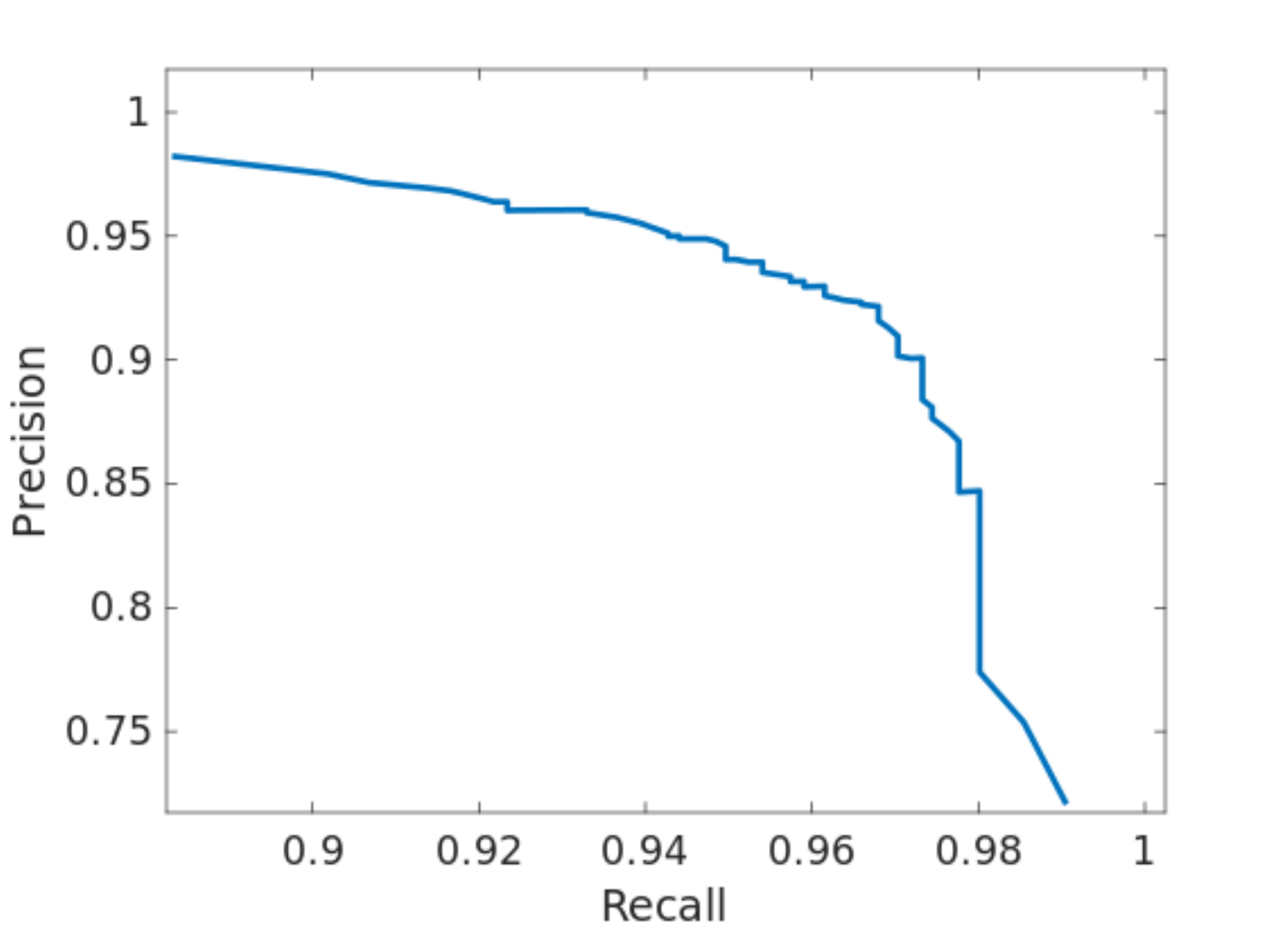}
\end{center}
   \caption{Precision-Recall curve of referee classifier.}
\label{fig:referee}
\end{figure}

Figure~\ref{fig:referee} shows precision-recall curve for referee classifier.

